\documentclass[sigconf]{acmart}

\AtBeginDocument{%
  \providecommand\BibTeX{{%
    \normalfont B\kern-0.5em{\scshape i\kern-0.25em b}\kern-0.8em\TeX}}}

\setcopyright{acmlicensed}
\copyrightyear{2018}
\acmYear{2018}
\acmDOI{XXXXXXX.XXXXXXX}

\acmISBN{978-1-4503-XXXX-X/18/06}

\usepackage{amssymb}
\usepackage{booktabs}
\usepackage{float}
\usepackage{tabularx}
\usepackage{bm}
\usepackage{mathrsfs}
\usepackage{amsmath}

\begin{document}

\title{UMSD:High Realism Motion Style Transfer via Unified Mamba-based Diffusion}


\author{Ziyun Qian}
\authornote{indicates the equal contribution.}
\affiliation{%
  \institution{Academy for Engineering and Technology, Fudan
University}
  \city{Shanghai}
  \country{China}}
\email{zyqian22@m.fudan.edu.cn}

\author{Zeyu Xiao}
\authornotemark[1] 
\affiliation{%
  \institution{Academy for Engineering and Technology, Fudan
University}
  \city{Shanghai}
  \country{China}}
\email{23210860082@m.fudan.edu.cn}

\author{Xingliang Jin}
\affiliation{%
  \institution{School of Computer Science and Technology, East China Normal University}
  \city{Shanghai}
  \country{China}}

\author{Dingkang Yang}
\authornotemark[2] 
\affiliation{%
  \institution{Academy for Engineering and Technology, Fudan
University}
  \city{Shanghai}
  \country{China}}
\email{dkyang20@fudan.edu.cn}

\author{Mingcheng Li}
\affiliation{%
  \institution{Academy for Engineering and Technology, Fudan
University}
  \city{Shanghai}
  \country{China}}

\author{Zhenyi Wu}
\affiliation{%
  \institution{Academy for Engineering and Technology, Fudan
University}
  \city{Shanghai}
  \country{China}}

\author{Dongliang Kou}
\affiliation{%
  \institution{Academy for Engineering and Technology, Fudan
University}
  \city{Shanghai}
  \country{China}}

\author{Peng Zhai}
\affiliation{%
  \institution{Academy for Engineering and Technology, Fudan
University}
  \city{Shanghai}
  \country{China}}

\author{Lihua Zhang}
\authornote{indicates the corresponding author.}
\affiliation{%
  \institution{Academy for Engineering and Technology, Fudan
University}
  \city{Shanghai}
  \country{China}}
\email{lihuazhang@fudan.edu.cn}

\renewcommand{\shortauthors}{Ziyun Qian, et al}
\renewcommand\footnotetextcopyrightpermission[1]{}

\begin{abstract}
Motion style transfer is a significant research direction in the field of computer vision, enabling virtual digital humans to rapidly switch between different styles of the same motion, thereby significantly enhancing the richness and realism of movements. It has been widely applied in multimedia scenarios such as films, games, and the metaverse. However, most existing methods adopt a two-stream structure, which tends to overlook the intrinsic relationship between content and style motions, leading to information loss and poor alignment. Moreover, when handling long-range motion sequences, these methods fail to effectively learn temporal dependencies, ultimately resulting in unnatural generated motions.  
To address these limitations, we propose a Unified Motion Style Diffusion (UMSD) framework, which simultaneously extracts features from both content and style motions and facilitates sufficient information interaction. Additionally, we introduce the Motion Style Mamba (MSM) denoiser, the first approach in the field of motion style transfer to leverage Mamba's powerful sequence modelling capability. Better capturing temporal relationships generates more coherent stylized motion sequences.  
Third, we design a diffusion-based content consistency loss and a style consistency loss to constrain the framework, ensuring that it inherits the content motion while effectively learning the characteristics of the style motion. Finally, extensive experiments demonstrate that our method outperforms state-of-the-art (SOTA) methods qualitatively and quantitatively, achieving more realistic and coherent motion style transfer.
\end{abstract}





\maketitle

\begin{figure*}[t]
  \centering
  \includegraphics[width=1.0\linewidth]{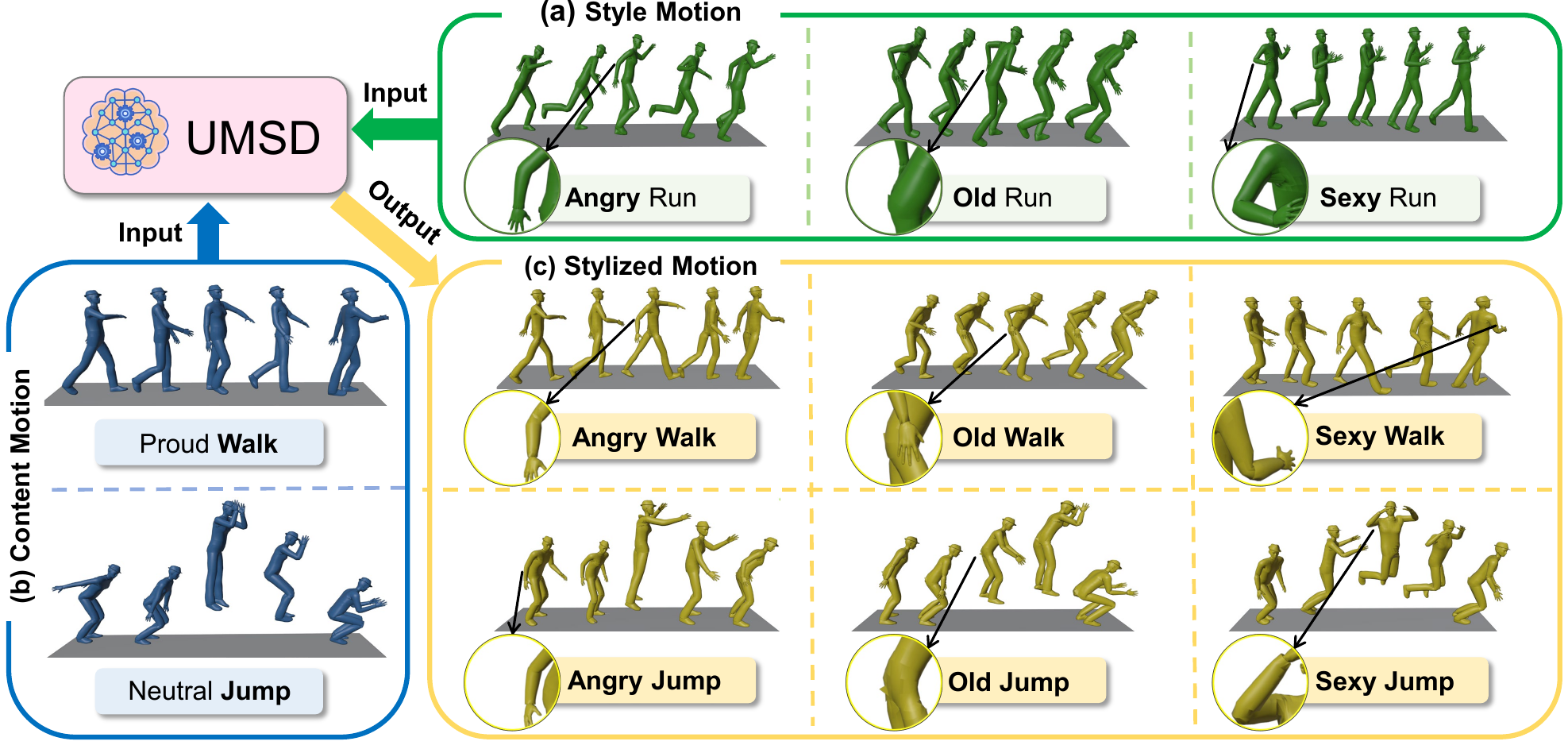}
  \caption{\textbf{Motion style transfer with UMSD.} Our method effectively transfers the style of the style motion to the content of the content motion, ensuring the prominence of the style while preserving the content of the motion to a great extent.
  }
\label{fig:teaser}
\end{figure*}

\section{Introduction}

\label{sec:intro}
Style is a crucial aspect of human motion, reflecting individual characteristics such as emotions, age, and health status \cite{style-definition}, which are essential for animating human characters and avatars. 
However, manually extracting these styles is both challenging and time-consuming \cite{manual-1,manual-2}.
Motion style transfer addresses this issue by transferring the style of a style motion onto the content of a content motion, resulting in a stylized motion that retains both characteristics (\textit{e.g.,} an angry walk in Figure \ref{fig:teaser}c). This technique enhances the diversity and realism of digital human motion, making it widely applicable in fields such as film production and game design.

\begin{figure}[t]
  \centering
  \includegraphics[width=1.0\linewidth]{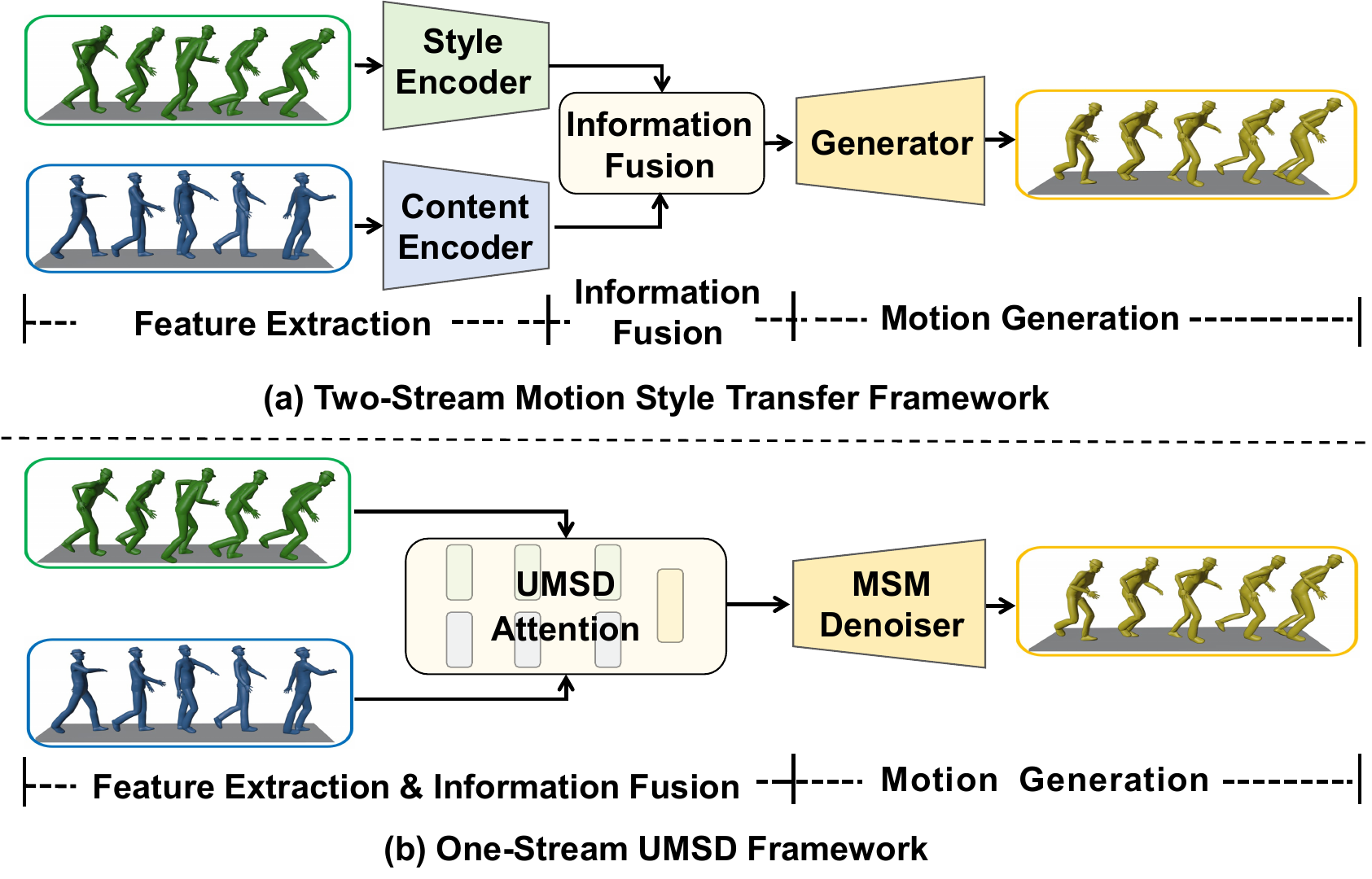}
  \vspace{-0.6cm}
  \caption{\textbf{Comparison between two-stream frameworks and our one-stream UMSD.} (a) Existing methods first extract content and style motion features separately, then perform feature fusion and motion generation. (b) In contrast, UMSD unifies feature extraction and information fusion, enabling direct generation of stylized motion.
  }
  \vspace{-0.5cm}
\label{UMSD Framework}
\end{figure}

However, achieving style characteristics while preserving motion content poses significant challenges. To address this issue, existing methods \cite{aberman2020unpaired,song2023finestyle,MCM,kim2024most} typically employ two separate encoders to independently extract features from content and style motions. The extracted information is fused, and a generative model produces stylized motion. Although this approach enables motion style transfer to some extent, the generated motion sequences still fall short in terms of naturalness and continuity for the following reasons: (1) Most current frameworks adopt a two-stream structure, with independent encoders for content and style motion, as shown in Figure \ref{UMSD Framework}a. This setup causes the encoders to overlook intrinsic connections between the two motion types, leading to information loss and poor alignment of features in high-dimensional space, ultimately degrading the quality of the generated output. (2) Current frameworks struggle to capture temporal relationships in long-range motion sequences, resulting in generated motion sequences that lack naturalness and coherence.

To address the above-mentioned issues, we propose a UMSD Framework, which employs a one-stream structure to extract features from content and style motions simultaneously. This unified approach enables effective information exchange, generates stylized motion, and overcomes the limitations of using separate encoders, as shown in Figure~\ref{UMSD Framework}b. Specifically, we introduce a novel UMSD Attention module that integrates cross-attention and self-attention mechanisms. The cross-attention mechanism enables information exchange between content and style motion features, enhancing their complementarity. Meanwhile, the self-attention mechanism independently processes each motion feature, capturing crucial local details and dependencies within the motion sequence.

To maintain long-range dependencies within the motion sequence, we first introduce the Mamba model \cite{mamba} for motion style transfer and propose the Motion Style Mamba (MSM) denoiser. MSM leverages the robust sequence modeling capability of State Space Models (SSM) to generate more coherent and natural motions by learning temporal dependencies. Also, we propose a diffusion-based style consistency loss and a diffusion-based content consistency loss, which constrain the UMSD framework to inherit motion content while learning motion style effectively. Extensive experiments on two benchmark datasets demonstrate that our method outperforms SOTA approaches.
Our contributions are as follows:

\begin{itemize}
\item We propose a novel UMSD framework employing a one-stream structure. This structure enables simultaneous feature extraction from content and style motion while facilitating extensive interaction between them.

\item We apply the Mamba model to the motion style transfer field for the first time and propose the MSM denoiser. This leverages SSM's strong sequential modeling capability to better preserve long-range dependencies in motion sequences.

\item We propose a Diffusion-based Content and Style Consistency Loss, which separately constrains the UMSD framework to more comprehensively retain input motion content while effectively learning style features.

\item We conduct extensive experiments to evaluate our framework, and the results show that the proposed UMSD framework outperforms SOTA methods in both qualitative and quantitative metrics.
\end{itemize}

\section{Related Work}
\subsection{Motion Style Transfer} 
Motion style transfer is an advanced and essential research area in computer vision. Early methods \cite{manual-2,manual-1} rely on handcrafted feature extraction to design various motion styles, which is inefficient and difficult to apply in practical contexts such as film and gaming. In recent years, new methods \cite{aberman2020unpaired,song2023finestyle,MCM,kim2024most,mu2023generative, zhang2024generative, raab2024single, zhong2024smoodi} based on deep learning techniques \cite{vaswani2017attention,cui2022mixformer,sun2024pixel,GAN,park2021diverse} have been proposed, effectively addressing the inefficiency of manual feature extraction. For instance, Aberman \textit{et al.} \cite{aberman2020unpaired} used unsupervised learning to transfer motion style by learning from a collection of style-labeled motions. Building on this, Finestyle \cite{song2023finestyle} and Most \cite{kim2024most}, respectively, designed a bidirectional interaction flow fusion module and an innovative motion style transformer, enabling effective learning of motion content and style feature transfer. MCM-LDM \cite{MCM} achieved high-quality motion style transfer by disentangling and finely integrating three key elements: motion trajectory, content, and style, ensuring the core content is preserved. 
However, most of these approaches adopted a two-stream structure, with two separate encoders extracting content and style motion features. This can cause the encoders to overlook intrinsic connections between the two motion types, leading to information loss and poor alignment in high-dimensional spaces.

\subsection{Diffusion Generative Models} 
Diffusion models are highly regarded for their exceptional performance in various research fields, including image generation \cite{img_generation_diffusion2024,diff-image2,diff-image3}, video generation \cite{video_diffusion_2024,diff-video2,diff-image3}, reinforcement learning \cite{reinforcement_diffusion_2024,diff-rein2}, and motion generation \cite{tevet2022MDM,Tseng_Castellon_Liu_2022,motion_generation_diffusion_2024,motion2_generation_diffusion_2024}.
For example, MDM~\cite{tevet2022MDM} utilized uses a transformer-based diffusion model for condition-guided motion generation. MLD~\cite{chen2023executing} introduced diffusion models in the latent space of a motion VAE, significantly improving high-fidelity motion generation. Alexanderson \textit{et al}.~\cite{alexanderson2023listen} explored diffusion models for audio-driven motion generation, demonstrating how auditory cues can guide motion synthesis.
However, in the above diffusion models, the denoiser often struggles to effectively learn the temporal dependencies in long sequences, which limits its performance in motion style transfer tasks.

\section{Methodology}
\noindent\textbf{Overview.} We propose a novel UMSD framework to achieve high naturalness in motion style transfer, as detailed in Section \ref{UMSD section}. First, we apply UMSD Attention to extract features from content and style motions and enable information exchange, as described in Section \ref{attention}. Then, we introduce an MSM denoiser, which generates more coherent stylized motion sequences (Section \ref{mamba section}). Section \ref{loss section} presents two loss functions to supervise content and style consistency.

\noindent\textbf{Pose Representation.} We categorize the motion sequences input into the UMSD framework into two types: content motion and style motion. Since the stylized motion output is strongly correlated with the input content and each motion is clearly defined by joint rotations (unit quaternions) \cite{2020TOG-ma}, we represent the content motion sequence as joint rotations $\boldsymbol m^{c,1:N}=\left\{\boldsymbol m^{c,i}\right\}_{i=1}^{N}\in \mathbb{R}^{4J\times N}$. Additionally, as style can be inferred from the relative motion of joint positions, we use joint positions to represent the style motion sequence $\boldsymbol n^{s,1:N}=\left\{\boldsymbol n^{s,i}\right\}_{i=1}^{N}\in \mathbb{R}^{3J\times N}$, where $J=21$ is the number of joints in the human skeleton \cite{aberman2020unpaired}, and $N$ represents the number of poses in a motion sequence. Here, $\boldsymbol m$ and $\boldsymbol n$ denote the motion content of content and style motion, respectively, and $c$ and $s$ indicate the motion style for content and style motion. The UMSD framework aims to learn the motion style $s$ while retaining the motion content $m$, thereby generating a stylized motion $\boldsymbol m^{s,1:N}=\left\{\boldsymbol m^{s,i}\right\}_{i=1}^{N}\in \mathbb{R}^{4J\times N}$ that combines both characteristics.

\begin{figure*}[t]
  \centering
  \vspace{-0.3cm}
  \includegraphics[width=1.0\linewidth]{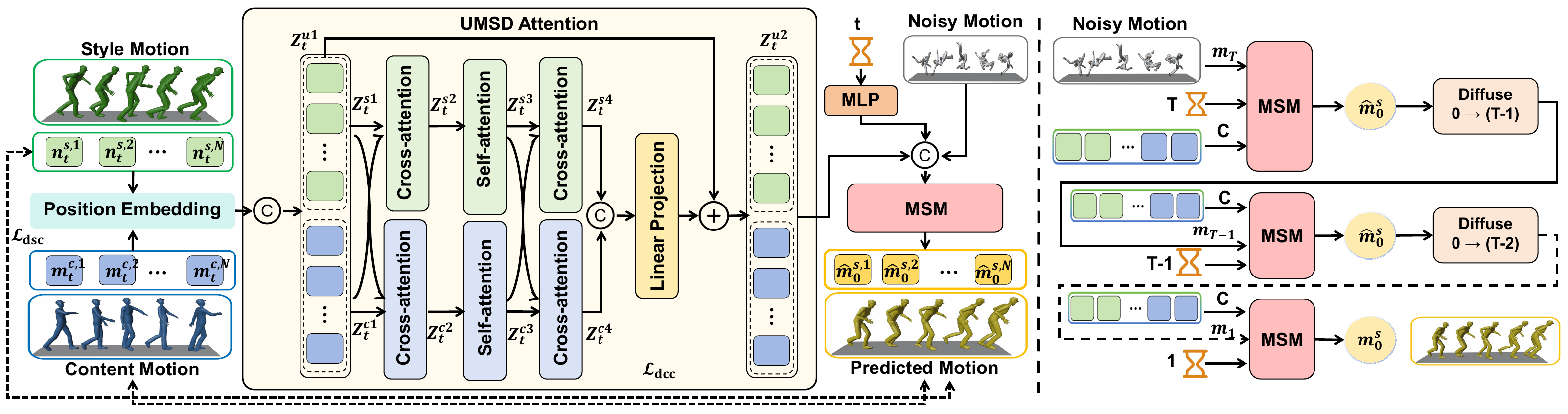}
  \vspace{-0.7cm}
  \caption{Framework overview. (Left) Overview of the Unified Motion Style Diffusion (UMSD) framework. The framework input is a noisy motion sequence $\boldsymbol m_t^{z,1:N}$ of length $N$ at noising step $\bm{t}$, with $z \in \{c, s\}$, conditional information $\boldsymbol{C}$, and $\bm{t}$ itself. Here, $\boldsymbol{C}$ represents the result of information interaction between content motion $\boldsymbol m_t^{c,1:N}$and style motion $\boldsymbol n_t^{s,1:N}$ through UMSD attention. A Motion Style Mamba (MSM) denoiser is then applied to predict the stylized motion $\hat {\boldsymbol m}_0^{s,1:N}$. (Right) Sampling MSM. Given the condition $\boldsymbol{C}$, noisy motion is sampled across the desired motion dimensions, iterating from $\bm{t}$ down to 1. At each step $\bm{t}$, the MSM predicts the clean stylized motion $\hat {\boldsymbol m}_0^{s,1:N}$, which is then diffused back to $\hat {\boldsymbol m}_{t-1}^{s,1:N}$.
}
\label{fig2}
\end{figure*}

\subsection{Unified Motion Style Diffusion Framework}
\label{UMSD section}
Existing frameworks \cite{aberman2020unpaired,song2023finestyle,kim2024most} for motion style transfer typically use a two-stream structure with separate encoders for content and style motion feature extraction. This structure often leads to information loss and misalignment in high-dimensional space. Moreover, when handling long-range motion sequences, these frameworks struggle to effectively model sequence dependencies and temporal relationships, resulting in stylized motions that lack natural flow and coherence. We propose a UMSD framework to address these issues, as illustrated in Figure \ref{fig2}.

Our UMSD framework is a one-stream structure based on the diffusion model \cite{motion_generation_diffusion_2024}. Taking the content motion $\boldsymbol m_{t}^{c,1}$ at noising step $\bm{t}$ as an example, diffusion is regarded as a Markov noising process. The motion sequence follows a forward noising process, $q(\boldsymbol m_{t}^{c,1:N}\mid \boldsymbol m_{t-1}^{c,1:N})$, where $\boldsymbol m_{0}^{c,1:N}$ is drawn from the data distribution. The forward noising process is defined as:
\begin{equation}
    q\big(\boldsymbol m_{t}^{c,1:N}\mid \boldsymbol m_{t-1}^{c,1:N}\big)=\mathcal N\big(\sqrt{\alpha_{t}}\boldsymbol m_{t-1}^{c,1:N},(1-\alpha_{t})I\big),
\end{equation}
where $\alpha_t \in (0,1)$ are constant hyperparameters. As $\alpha_t$ approaches 0, we approximate $\boldsymbol m_{T}^{c,1:N} \sim \mathcal N (0,1)$. We use $T = 1000$ timesteps. The forward noising process for the style motion $\boldsymbol n_{t}^{s,1:N}$ follows the same procedure. From this point, we denote the full-length sequences of content motion, style motion, and stylized motion at noising step $\bm{t}$ as $\boldsymbol m_t^c$, $\boldsymbol n_t^s$, and $\boldsymbol m_t^s$, respectively.

\subsection{UMSD Attention}
\label{attention}

During motion style transfer, the accurate encoding of both style and content is essential. Existing methods~\cite{aberman2020unpaired,song2023finestyle,MCM,kim2024most} typically encode them separately, which causes the encoders to overlook the intrinsic connections between the two motion types. We propose UMSD Attention, which integrates cross-attention and self-attention mechanisms to extract features from both content and style motions, facilitating comprehensive information exchange, as shown in Figure \ref{fig2}.

To begin with, we concatenate the position-encoded content motion sequence $\boldsymbol m_t^c$ and style motion sequence $\boldsymbol n_t^s$ at noising step $\bm{t}$, resulting in the unified sequence $\boldsymbol Z_{t}^{u1}$, defined as $\boldsymbol Z_{t}^{u1}=[\boldsymbol Z_{t}^{c1};\boldsymbol Z_{t}^{s1}]$. Here, each pose in the motion sequence represents a token. We perform feature extraction and information fusion in three stages, using $\boldsymbol Z_{t}^{s\it{i}}$ with $i \in \{1,2,3,4\}$ as query embeddings in the following illustration. In the first stage, we employ a cross-attention mechanism to facilitate information exchange between content and style motion features, enhancing their complementarity. This process is expressed as follows:
\begin{equation}
    \boldsymbol Z_{t}^{s2}=\mathrm{softmax}\left(\frac{\boldsymbol Q_{s1}\boldsymbol K_{c1}^{T}}{\sqrt{d}}\right)\boldsymbol V_{c1}.
\end{equation}
The query $\boldsymbol Q_{s\it i}$ is generated by applying a linear projection to $\boldsymbol Z_t^{s \it i}$. At the same time, $\boldsymbol K_{c \it i}^{T}$ and $\boldsymbol V_{c \it i}$ are produced by linearly projecting $\boldsymbol Z_t^{c \it i}$ to obtain keys and values, respectively, where $i \in \{1,2,3,4\}$ and $d$ represents the dimension of the key. We then apply a self-attention mechanism to independently process the content and style motion features, capturing critical local details and dependencies across motion sequences, yielding $\boldsymbol Z_{t}^{s3}$. In the third stage, we use a cross-attention mechanism to establish deeper connections between the content and style motions, which helps the final stylized motion better integrate both feature types. The following formula represents this process:
\begin{equation}
    \boldsymbol Z_{t}^{s3}=\mathrm{softmax}\left(\frac{\boldsymbol Q_{s2}\boldsymbol K_{s2}^{T}}{\sqrt{d}}\right)\boldsymbol V_{s2},
\end{equation}
\begin{equation}
    \boldsymbol Z_{t}^{s4}=\mathrm{softmax}\left(\frac{\boldsymbol Q_{s3}\boldsymbol K_{c3}^{T}}{\sqrt{d}}\right)\boldsymbol V_{c3}.
\end{equation}
$\boldsymbol K_{s\it i}^{T}$ and $\boldsymbol V_{s\it i}$ represent the keys and values generated from $\boldsymbol Z_{t}^{s\it i}$, with $i \in \{1,2,3,4\}$, through linear projection. The process for obtaining query embeddings $\boldsymbol Z_t^{c \it i}$ follows the same steps, resulting in $\boldsymbol Z_t^{c4}$. By concatenating it with $\boldsymbol Z_t^{s4}$, we obtain $\boldsymbol Z_t^{u2}$ as $\boldsymbol Z_t^{u2} = [\boldsymbol Z_t^{c4};\boldsymbol Z_t^{s4}]$. Subsequently, the following operation is applied to yield the output $\boldsymbol Z_t^{out}$ for the UMSD attention:
\begin{equation}
\boldsymbol{Z}_{t}^{out}=\boldsymbol Z_{t}^{u1}\oplus \text{LN}(\boldsymbol Z_{t}^{u2}),
\end{equation}
where $\oplus$ denotes the matrix addition, and $\text{LN}(\cdot)$ is a linear layer.

Through this interactive attention mechanism, UMSD attention effectively encodes both style and content, allowing them to mutually reinforce each other. This enables the style transfer results to not only faithfully express the style but also retain the content to a significant extent.

\subsection{Motion Style Mamba Denoiser}
\label{mamba section}
\begin{figure}[t]
  \centering
\includegraphics[width=1.0\linewidth]{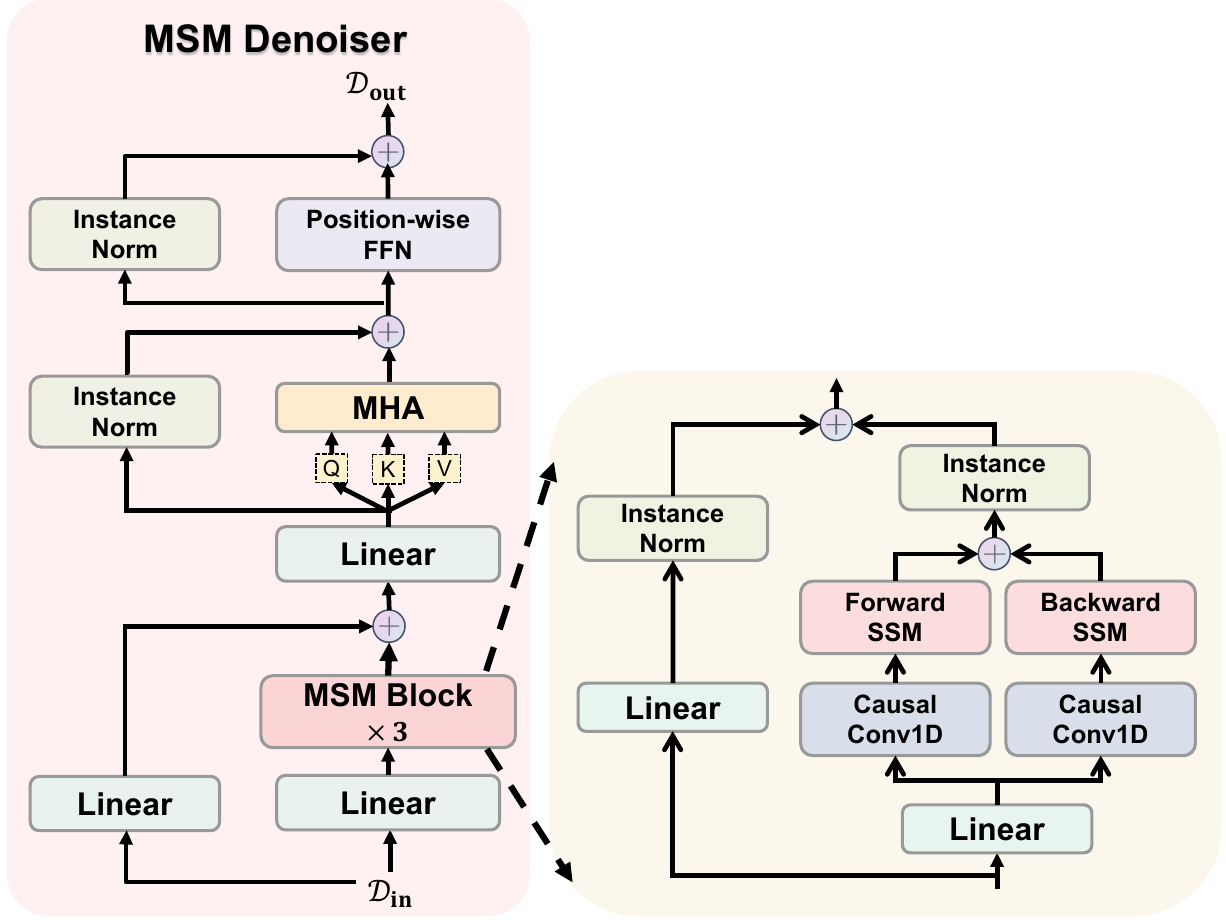}
  \caption{\textbf{Architecture of MSM denoiser.} 
  }
\label{fig-MSM}
\end{figure}

After employing diffusion-based models \cite{DDPM_ho2020}, we observe that existing denoisers, such as U-Net \cite{unet} and transformers \cite{liu2023itransformer, vaswani2017attention}, struggle to capture temporal relationships in long-range motion sequences effectively \cite{tevet2022MDM}. This limitation leads to generated motion sequences that lack natural continuity. To address this issue, we draw inspiration from the Mamba model \cite{mamba} and apply it for the first time in the motion style transfer field, proposing the Motion Style Mamba (MSM) denoiser. MSM leverages the powerful sequence modelling capability of the State Space Model (SSM) to capture temporal information in motion sequences better, preserving long-term dependencies within them. The structural diagram is shown in Figure \ref{fig-MSM}.
Before entering the MSM denoiser, the motion sequence $\boldsymbol{m}_z^t$ at noising step $\bm{t}$, with $\boldsymbol{z} \in \{c, s\}$, undergoes the following processing:
\begin{equation}
    \mathcal{D}_{in}=\text{concat}(\boldsymbol m_{t}^{z},\text{MLP}(T),\boldsymbol Z_{t}^{out}),
\end{equation}
where $\text{concat}(\cdot)$ represents concatenation, $D_{in}$ represents the input to the MSM denoiser. The \text{MLP} consists of two linear layers and an activation layer, projecting the timestep $\bm{t}$ into a continuous vector space to form a latent vector optimized for MSM processing. 

The MSM block, the core of the MSM denoiser, leverages the long-range sequential modeling strengths of the SSM to map the timestep $\bm{t}$ into content and style motion sequences, thereby extracting temporal information while preserving long-range dependencies in the motion sequence. Its structure is as follows:
\begin{equation}
\begin{gathered}
\mathcal{D}^0=\text{LN}\left(\mathcal{D}_{\text {in }}\right), \\
\mathcal{D}_r^i=\text{IN}\left(\Phi^{+}\left(\mu\left(\text{LN}\left(\mathcal{D}^{i-1}\right)\right)\right)+\Phi^{-}\left(\mu\left(\text{LN}\left(\mathcal{D}^{i-1}\right)\right)\right)\right), \\
\mathcal{D}^i=\text{IN}\left(\text{LN}\left(\mathcal{D}^{i-1}\right)\right)+\mathcal{D}_r^i, \\
\mathcal{D}^{\text {res }}=\text{LN}\left(\mathcal{D}_{\text {in }}\right)+\mathcal{D}^3,
\end{gathered}
\end{equation}
where $\text{LN}(\cdot)$ denotes a linear layer, $\text{IN}(\cdot)$ represents an InstanceNorm layer, $\Phi^{+}(\cdot)$ refers to forward SSM, and $\Phi^{-}(\cdot)$ to backward SSM. The original Mamba model, designed for 1-D sequences, is unsuitable for motion style transfer tasks requiring spatial awareness, so we adopt bidirectional sequence modeling here. $\mu(\cdot)$ denotes a Causal Conv1D layer used for feature extraction, $\mathcal{D}_r^i$ and $\mathcal{D}^i$ represent the intermediate result and final output of the right branch in the $i-th$ iteration of the MSM Block respectively, where $i \in \{1,2,3\}$. Notably, $\mathcal{D}^{0}$ represents the input to the MSM Block, while $\mathcal{D}^{res}$ is the output of the residual network containing the MSM Block.

Following the MSM Block structure, we integrate the sequential modelling strengths of SSM with the contextual awareness of the attention mechanism, enhancing our ability to capture fine-grained temporal changes at each timestep of the motion sequence, thereby generating more naturally stylized motion:

\begin{equation}
    \sigma=\text{IN}(\text{LN}(\mathcal{D}^{res}))+\text{MHA}({\text{LN}}(\mathcal{D}^{res})),
\end{equation}
where $\text{MHA}(\cdot)$ represents Multi-Head Attention and $\sigma$ denotes the output of the residual network containing \text{MHA}. The output of the final MSM denoiser $\mathcal{D}_{out}$, is expressed by the following equation:
\begin{equation}
    \mathcal{D}_{out}=\text{FFN}(\sigma)+\text{IN}(\sigma),
\end{equation}
where $\text{FFN}(\cdot)$ represents Feed-Forward Network.
With the integration of the Mamba model, our MSM denoiser is better equipped to capture the temporal information of long motions, enabling the generation of more coherent and natural stylized motion sequences.

\subsection{Training Objectives}
\label{loss section}

Our objective is to use the output of UMSD Attention as the condition $c$ for the diffusion model, allowing the framework to learn the motion style $s$ while retaining the motion content $ \bm{m}$, thus generating stylized motion $\bm{m}_t^s$. However, existing diffusion-based motion style transfer methods \cite{MCM} rely solely on a simple reconstruction loss, which results in insufficient preservation of motion content and style coherence. To address this, we propose the Diffusion-based content consistency loss (\textit{i.e.,} Eq. \ref{dcc}) and style consistency loss (Eq. \ref{dsc}), which ensure both accurate retention of content and faithful transfer of style throughout the motion generation process.


\noindent\textbf{Diffusion-based Content Consistency Loss.} When the input content motion sequence $\boldsymbol{m}_t^c$ and style motion sequence $\boldsymbol{n}_t^s$ share the same style, i.e., $c = s$, the generated stylized motion $\boldsymbol{m}_t^s$ should closely resemble the content motion $\boldsymbol{m}_t^c$, regardless of the content of style motion $\boldsymbol{n}_t^s$. Based on this observation, we model the distribution $p(\boldsymbol{m}_0^c|\mathbf{C})$ as a reverse diffusion process that iteratively denoises $\boldsymbol{m}_T^c$. Instead of predicting noise $\epsilon_t$ following the formula in DDPM \cite{DDPM_ho2020}, we adopt an equivalent approach from Ramesh \textit{et al.} \cite{ramesh2022hierarchical}, directly predicting the motion itself. Specifically, $\hat{\boldsymbol{m}}_0^s = \hat{\boldsymbol{m}}_0^c = \text{MSM}(\boldsymbol{m}_t^c, \bm t, \boldsymbol{C}) = \text{MSM}(\boldsymbol{m}_t^c, \bm t, \text{U}(\boldsymbol{m}_0^c, \boldsymbol{n}_0^s))$ (see Figure \ref{fig2}, right). The diffusion-based content consistency loss is expressed as:
\begin{equation}
\label{dcc}
    \mathcal{L}_{\mathrm{dcc}} =\mathbb{E}_{\bm m_{0}^{c},\boldsymbol n_{0}^{s}\sim\mathcal{M}}\|\text{MSM}(\boldsymbol m_{t}^{c},\bm{t},\text{U}(\boldsymbol m_{0}^{c},\boldsymbol n_{0}^{s}))-\boldsymbol m_{0}^{c}\|_{1},
\end{equation}
where $\mathcal{M}$ denotes the dataset, ${\text{MSM}}(\cdot)$ represents the Motion Style Mamba denoiser, and $\text{U}(\cdot)$ denotes the UMSD Attention module. Our loss function differs fundamentally from the content consistency loss used in other methods \cite{2020TOG-ma, park2021diverse} in two main aspects: (1) it is based on a diffusion model, enabling control over the noise addition process to motion via the timestep $\bm{t}$; (2) the condition in our loss function is the result of fusing content and style motion features, allowing the framework to learn both features better. 

\noindent\textbf{Diffusion-based Style Consistency Loss.} We adopt a similar approach as above. If content motion sequence $\boldsymbol{m}_t^c$ and style motion sequence $\boldsymbol{n}_t^s$ share the same content, i.e., $\boldsymbol{m}=\boldsymbol{n}$, the stylized motion $\boldsymbol m_t^s$ should ideally be as close as possible to the style motion $\boldsymbol n_t^s$. We use the MSM denoiser to directly predict the motion itself, i.e., $\hat {\boldsymbol m}_{0}^{s}=\hat {\boldsymbol n}_{0}^{s}=\text{MSM}(\boldsymbol n_{t}^{s},\bm{t},\boldsymbol C)=\text{MSM}(\boldsymbol n_{t}^{s},\bm{t},\text{U}(\boldsymbol m_{0}^{c},\boldsymbol n_{0}^{s}))$. The diffusion-based style consistency loss is expressed as follows:
\begin{equation}
\label{dsc}
    \mathcal{L}_{\mathrm{dsc}} =\mathbb{E}_{\boldsymbol m_{0}^{c},\boldsymbol n_{0}^{s}\sim\mathcal{M}}\|\text{MSM}(\boldsymbol n_{t}^{s},\bm{t},\text{U}(\boldsymbol m_{0}^{c},\boldsymbol n_{0}^{s}))-\boldsymbol n_{0}^{s}\|_{1}.
\end{equation}

Additionally, we adopt three existing geometric losses, $\mathcal{L}_{\mathrm{pos}}$, $\mathcal{L}_{\text {foot}}$, and $\mathcal{L}_{\text {vel }}$, which control positions, foot contact, and velocities, respectively \cite{shi2020motionet,tevet2022MDM,Tseng_Castellon_Liu_2022}. Our total training loss function is a combination of the above five losses:
\begin{equation}
\mathcal{L}_{\text {total }}=\mathcal{L}_{\mathrm{dcc}}+\mathcal{L}_{\mathrm{dsc}}+\mathcal{L}_{\mathrm{pos}}+\mathcal{L}_{\text {vel }}+\mathcal{L}_{\text {foot}}.
\end{equation}

\begin{figure*}[t]
  \centering
  \includegraphics[width=1.0\linewidth]{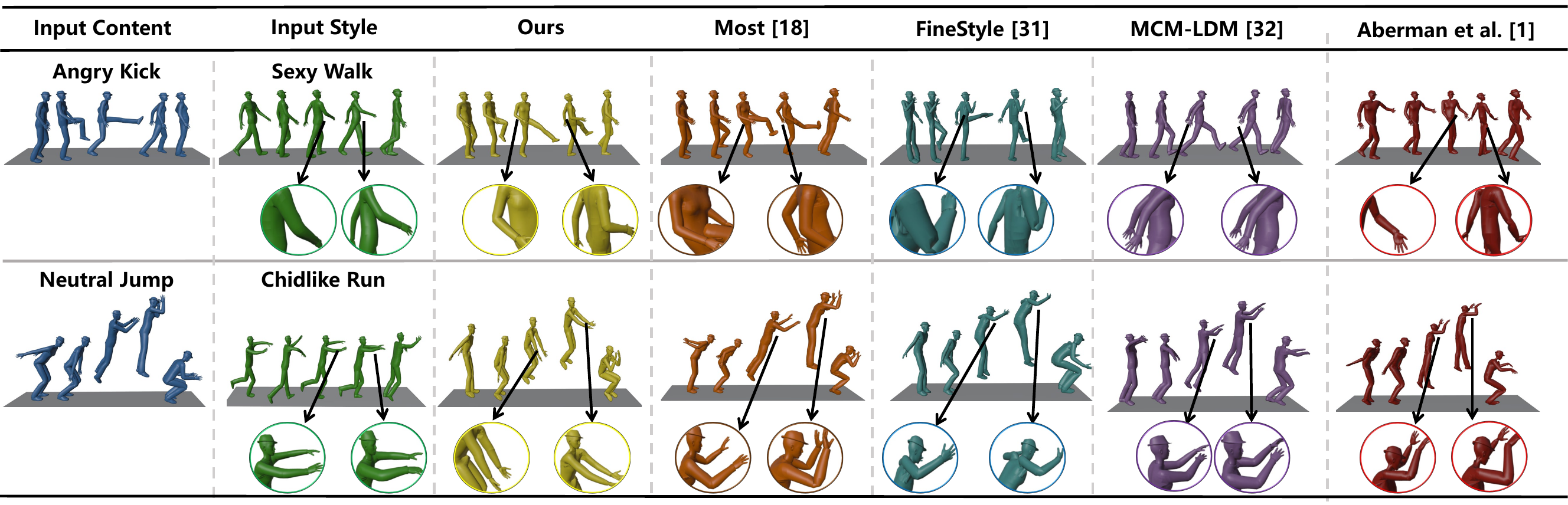}
  \caption{\textbf{Qualitative comparison of short sequences.} We provide two sets of cases comparing the style transfer effects with SOTA methods \cite{aberman2020unpaired,MCM,song2023finestyle,kim2024most}. We zoom in on critical areas that reflect style characteristics for a more intuitive assessment. The results indicate that our UMSD framework performs better.
  }
\label{fig4}
\end{figure*}

\begin{figure*}[t]
  \centering
  \includegraphics[width=1.0\linewidth]{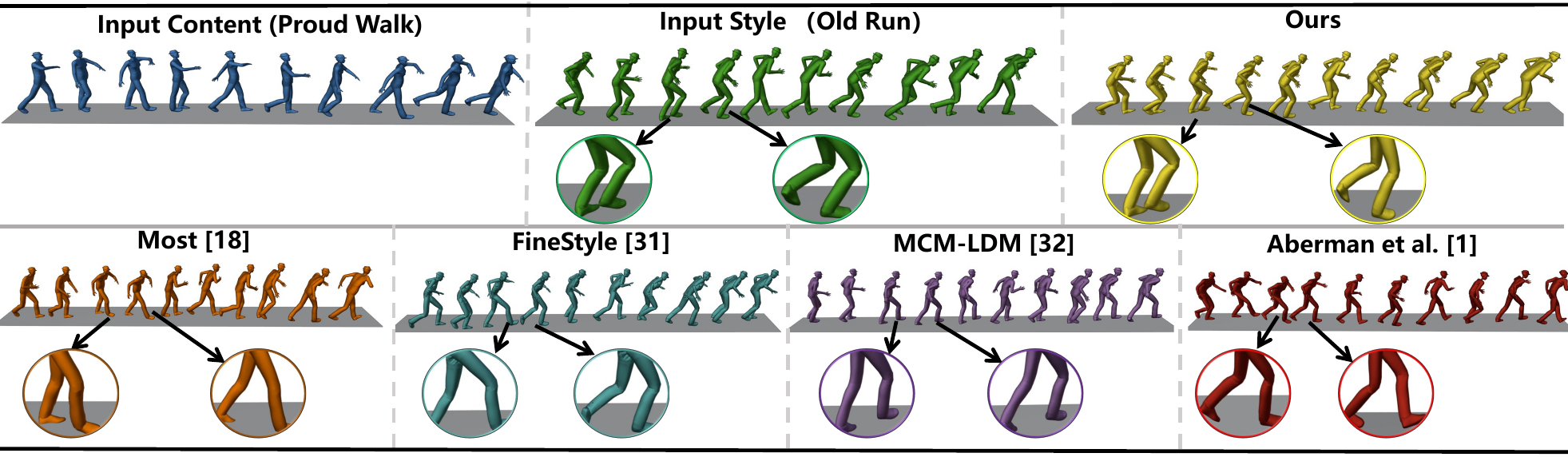}
  \caption{\textbf{Qualitative comparison of long sequences}. We select two long-sequence samples from the Xia dataset \cite{xia2015realtime} for motion style transfer. The results demonstrate that our UMSD framework outperforms SOTA methods \cite{aberman2020unpaired,song2023finestyle,kim2024most,MCM}.
  }
\label{fig-long}
\end{figure*}

\section{Experiments}

In this section, we conduct a series of experiments to evaluate the effectiveness of the UMSD framework. First, we provide the details of the implementation of the experiments. Then, we perform both quantitative and qualitative comparisons between UMSD and state-of-the-art (SOTA) methods on two datasets. In the qualitative comparison, we separately assess the model's motion style transfer capability for both short and long motion sequences.  

Additionally, to further evaluate UMSD's generalization ability, we test style transfer using previously unseen styles from the training dataset. In the ablation study, we demonstrate the effectiveness of each module in the UMSD framework. A user study is also conducted to compare our method's performance with SOTA methods from a more intuitive perspective. More experimental results can be found in the technical appendix.

\subsection{Implementation Details}
We train and test our model based on the Xia dataset \cite{xia2015realtime} and BFA dataset \cite{aberman2020unpaired}. 
We reduce the original 120fps motion data to 60fps and obtain approximately 1500 motion sequences in total. Our framework is implemented in PyTorch and trains on NVIDIA A800 GPUs, with a learning rate of $e^{-6}$, using the AdamW optimizer \cite{opt}. The training process takes about 10 hours.

\subsection{Quantitative Evaluation}
\label{定量比较章节}

We employ the metrics of FMD, KMD, Diversity, CRA, and SRA \cite{raab2023modi,song2023finestyle,MCM,puzzle} to evaluate our framework quantitatively. The first two metrics are variants of Fréchet Inception Distance (FID) \cite{FID} and Kernel Inception Distance (KID) \cite{KID}, respectively, measuring the distribution discrepancy between generated and real motion sequences. Diversity quantifies the diversity of generated motions. We train our feature extractor to compute the values of these three metrics. Lower FMD and KMD values indicate that the generated motions are closer to real motions, implying higher generation quality. Conversely, higher Diversity values reflect more extraordinary richness in the generated motions. We compute the content preservation degree and style recognition accuracy of the generated motions using our self-trained content classifier and style classifier for the CRA and SRA metrics. Higher values of these two metrics indicate better quality of the generated stylized motions.

We conduct quantitative comparisons on two mainstream datasets, the Xia \cite{xia2015realtime} and BFA \cite{aberman2020unpaired} datasets. The results are presented in Tables \ref{tab-Xia} and \ref{tab-BFA}. It can be observed that our proposed method outperforms SOTA methods \cite{aberman2020unpaired,MCM,song2023finestyle,kim2024most} on most metrics in the Xia dataset \cite{xia2015realtime}. This superior performance primarily stems from our UMSD framework's ability to facilitate sufficient information interaction between content and style motions. Although our method lags behind Aberman \textit{et al.}'s method \cite{aberman2020unpaired} regarding Diversity on the Xia dataset, this is mainly because our model architecture prioritizes generating more natural and realistic stylized motions rather than maximizing motion diversity.

Our UMSD framework achieves even more outstanding results on the long-sequence BFA dataset \cite{aberman2020unpaired}, surpassing SOTA methods across all metrics. This significant improvement primarily benefits from our proposed MSM denoiser, which leverages the State Space Model's (SSM) powerful sequence modelling capability to better capture temporal dependencies in motion sequences. The MSM denoiser generates more natural and coherent motion sequences by effectively modelling long-range temporal relationships.


\subsection{Qualitative Evaluation}
We qualitatively compare the visual effects of motion style transfer between our proposed UMSD framework and SOTA methods \cite{aberman2020unpaired,MCM,song2023finestyle,kim2024most} from three aspects: style expressiveness, content preservation, and motion realism. The content motion and style motion used in the experiments are sourced from the Xia dataset \cite{xia2015realtime}. 
Ideally, the model can learn the style of the style motion while preserving the content of the content motion, thereby generating stylized motion that combines the characteristics of both.

\noindent \textbf{Short Sequence Evaluation.} As shown in Figure \ref{fig4}, we conduct two sets of motion style transfer experiments with motions fewer than 190 frames to fairly compare our UMSD with other methods trained on short motion clips. The results demonstrate that our UMSD framework generates better visual effects. 
For example, in the first row, we transfer the sexy style to the kick motion. In the input style motion, the right arm is at a right angle, and the left arm hangs straight down. Our generated kick motion maintains these characteristics. Other methods \cite{aberman2020unpaired,song2023finestyle,kim2024most,MCM}, however, fail to capture this feature and produce awkward motions. 
This is due to our use of a one-stream architecture, where style and content information exchanged during encoding. Thus, our style transfer achieves both effective style representation and content preservation.


\noindent \textbf{Long Sequence Evaluation.} To evaluate the ability of our framework to generate long motion sequences, we select two samples from the Xia dataset \cite{xia2015realtime}, each with over 190 frames, as long-sequence content and style motions for qualitative comparison.
As shown in Figure \ref{fig-long}, when transferring the old style to the walk motion, the motion generated by our method shows more pronounced leg curvature, a smaller gap between the thighs, and greater consistency with the input style compared to SOTA methods \cite{aberman2020unpaired,MCM,song2023finestyle,kim2024most}. It also better captures the characteristics of the old style, demonstrating stronger style expressiveness. This superior performance is primarily attributed to the powerful sequence modelling capability of the SSM module, which more effectively captures the temporal information of the motion sequence, thereby preserving long-term temporal dependencies within the sequence.

\begin{figure*}[t]
  \centering
  \includegraphics[width=1.0\linewidth]{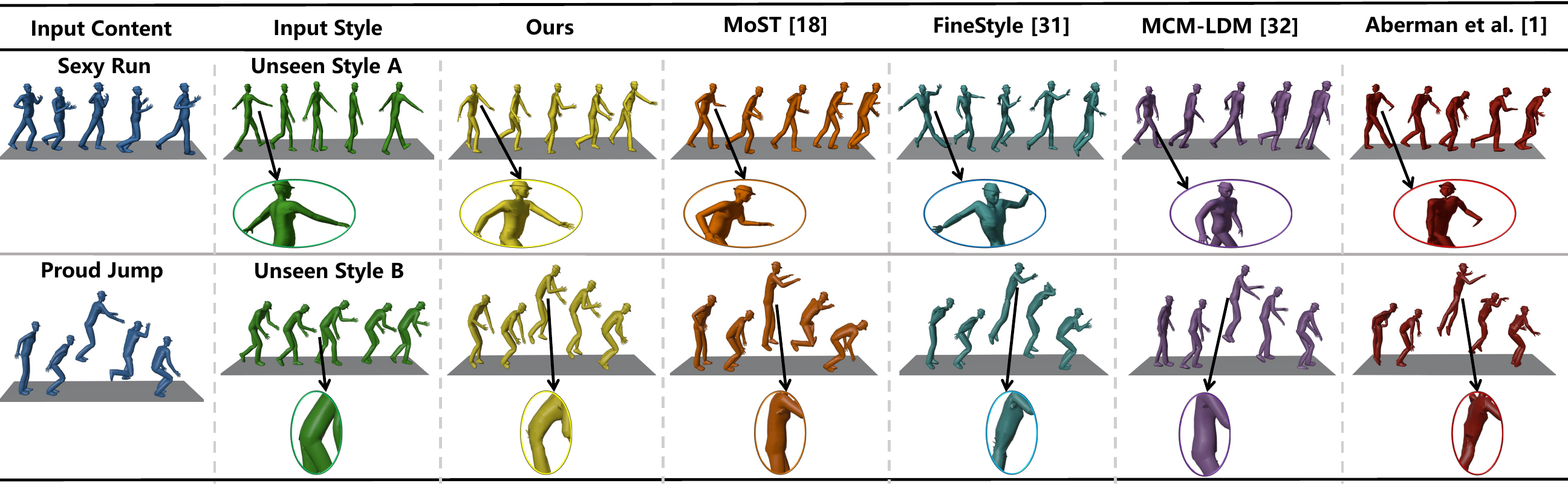}
  \caption{\textbf{Generalization evaluation.} We select a unseen style from the BFA dataset \cite{aberman2020unpaired}, which are not included in the training process, and transfer them to the sexy run and proud jump motions. The stylized motions are presented in the last four columns. Our method achieves better results than SOTA methods.
  }
\label{fig5}
\end{figure*}

\subsection{Generalizability Evaluation}
Our model can extract styles from arbitrary motion clips. However, in practical applications, motion style transfer models will likely encounter style categories outside the training dataset. Whether the model can still transfer styles from unseen styles determines its generalizability and practical utility.  

To compare the generalizability of our proposed UMSD framework with other methods, we select two styles, ``happy'' and ``sneaky'', from the BFA dataset \cite{aberman2020unpaired} (which is not involved in training) as unseen style A and unseen style B (second column), respectively, for testing. We then transfer these two unseen styles to two content motions, ``run'' and ``jump'' (first column), and compare the stylized motions generated by different methods.

The results are shown in Figure \ref{fig5}, where we zoom in on key body parts that reflect stylistic characteristics. In the first row, our UMSD framework successfully captures the arm-spreading motion when handling unseen style A. In contrast, other SOTA methods \cite{aberman2020unpaired,MCM,song2023finestyle,kim2024most} generate unnatural movements and fail to transfer the style effectively. In the second row, we transfer unseen style B to a proud jump motion—our method preserves the original jumping motion while accurately adopting the spinal curvature characteristic of the unseen style, while other approaches fail to achieve this.  
The generalizability comparison shows that our UMSD exhibits stronger generalization capability and practical utility, making it more effective for real-world applications in multimedia fields such as film production, game design, and virtual reality.

\subsection{Ablation Study}
Here, we conduct ablation experiments on our key components in Table~\ref{tab-loss-ablation}, including UMSD Attention, SSM, MHA, loss functions $\mathcal{L}_{\mathrm{dcc}}$ and $\mathcal{L}_{\mathrm{dsc}}$. In Table~\ref{tab_model_ablation}, we perform a module comparison of our MSM with alternative structures, such as STGCN and iTransformer.

\noindent \textbf{Importance of UMSD Attention and Loss Functions.}
To evaluate the effectiveness of the UMSD attention module, we remove it and instead use two independent encoders to extract content and style motion features separately. These features are then fused to generate stylized motion. As shown in Table \ref{tab-loss-ablation}, all evaluation metrics exhibit degradation, demonstrating that our one-stream structure enables more comprehensive information exchange, thereby reducing misalignment in high-dimensional space. We also conduct ablation experiments on the loss functions, confirming their effectiveness in constraining the model to capture motion content and style features better.


\noindent \textbf{Importance of the MSM Denoiser.}
We conduct ablation studies on the MSM Denoiser to ensure a fair comparison. Specifically, we replace the MSM module with STGCN \cite{STG} and iTransformer \cite{liu2023itransformer}, respectively, retrain the framework, and perform comparative experiments. As shown in Table \ref{tab_model_ablation}, the performance deteriorates significantly after substituting the MSM module with either of these alternatives. Our MSM module surpasses both alternatives across all quantitative evaluation metrics.  

These results demonstrate that STGCN \cite{STG} and iTransformer \cite{liu2023itransformer} structures are substantially inferior to our model in capturing the global temporal dynamics of motion sequences, particularly in maintaining long-range temporal dependencies. Furthermore, this comparison highlights our method's superior sequence modelling capability, which more effectively captures temporal information in motion sequences and generates more coherent and natural stylized motion sequences.


Additionally, we conduct ablation studies on the MSM Denoiser’s internal SSM and MHA structures in Table \ref{tab-loss-ablation}. Removing either results in degraded performance, demonstrating that the combination of SSM’s sequential modelling strengths and the contextual awareness of the attention mechanism effectively captures fine-grained variations in motion sequences, leading to optimal results.

\subsection{User Study}
In addition to qualitative and quantitative comparisons, we conduct a user study to evaluate various methods' motion style transfer results. We convert the generated stylized motions into videos and include them in a questionnaire. 50 volunteers assess the motions based on three criteria: (1) Content Preservation (CP): Does the generated motion retain the content of the content motion? (2) Style Expressiveness (SE): Does the generated motion capture the style characteristics? (3) Motion Realism (MR): Is the generated motion realistic and natural? Volunteers rate each criterion from 1 (not achieved) to 10 (fully achieved).

After collecting all the questionnaires, we set the confidence level of 95\% and calculate the average scores from the 50 volunteers. The results, shown in Table \ref{tab-user study}, indicate that our method achieves the highest CP, SE, and MR scores, with particularly outstanding performance in MR. Additionally, we perform an ANOVA test to examine the significance of these differences. The overall ANOVA establishes considerable distinctions among CP ($\mathit{F}$=14.847, $\mathit{p}$\textless0.01), SE ($\mathit{F}$=8.299, $\mathit{p}$\textless0.01), and MR ($\mathit{F}$=32.313, $\mathit{p}$\textless0.01). The post-hoc analysis suggests that our UMSD framework scores significantly higher than other methods \cite{aberman2020unpaired,MCM,song2023finestyle,kim2024most} across all three metrics (all $\mathit{p}$\textless0.01). 


These results further demonstrate the superior performance of our framework. The UMSD framework achieves outstanding results primarily through its one-stream structure, which extracts features from content and style motions while enabling effective information interaction. Additionally, our proposed MSM denoiser utilizes the powerful sequence modelling capability of state space models to generate more coherent and natural motions, further enhancing the realism of stylized motions.




\section{Conclusion}
In this work, we propose the UMSD framework, which employs a one-stream structure to extract features from content and style motion, enabling comprehensive information interaction and avoiding the limitations of using two separate encoders. We also introduce the MSM denoiser, which, for the first time in motion style transfer, leverages the robust sequential modelling capacity of SSM to learn temporal information and enhance motion coherence. Additionally, we propose two loss functions to guide model training. Finally, extensive experiments on two benchmark datasets demonstrate that our method surpasses SOTA approaches. We hope our work inspires further research in this field, leading to more practical applications in real-world scenarios.

\noindent \textbf{Future Work.} The field of motion style transfer presents several promising directions for future research, including cross-modal style transfer and style transfer under physical constraints. The supplementary materials provide detailed discussions.

\bibliographystyle{ACM-Reference-Format}
\bibliography{sample-base_2025MM}










\end{document}